%% file: acl_latex.tex
\pdfoutput=1

\documentclass[11pt]{article}

\usepackage[dvipsnames]{xcolor}
\usepackage[]{acl}

\usepackage{times}
\usepackage{latexsym}
\usepackage{url}

\usepackage{amsmath}
\usepackage{amssymb}
\usepackage[scaled=0.8]{beramono}
\usepackage{booktabs}
\usepackage{colortbl}
\usepackage{graphicx}
\usepackage{enumitem}
\usepackage{multicol}
\usepackage{multirow}
\usepackage{tcolorbox}
\usepackage{todonotes}
\usepackage{xspace}
\usepackage{pifont}
\usepackage{tabularx}

\newcommand{\dataset}{\textsc{PeopleProfiles}\xspace}
\newcommand{\stoe}{\texttt{\textbf{ENTITY}}\xspace}
\newcommand{\stob}{\texttt{\textbf{BODY}}\xspace}
\newcommand{\btol}{\texttt{\textbf{LEAD}}\xspace}

\newcommand{\fever}{\textbf{F}\xspace}
\newcommand{\wice}{\textbf{W}\xspace}
\newcommand{\vitaminc}{\textbf{VC}\xspace}
\newcommand{\wikifactcheck}{\textbf{WFC}\xspace}

\usepackage[T1]{fontenc}

\usepackage[utf8]{inputenc}

\usepackage{microtype}

\title{How Grounded is Wikipedia? \\ A Study on Structured Evidential Support and Retrieval}

\author{William Walden\thanks{~~Equal Contribution.} \quad Kathryn Ricci\footnotemark[1] \quad Miriam Wanner \quad Zhengping Jiang \\
\textbf{Chandler May} \quad \textbf{Rongkun Zhou} \quad\textbf{Benjamin Van Durme} \\
  Johns Hopkins University \\
  \texttt{\small{\{wwalden1, kricci2\}@jhu.edu}}}

\begin{document}
\maketitle
\begin{abstract}
\input{sections/00-abstract}
\end{abstract}

\section{Introduction}
\label{sec:introduction}
\input{sections/01-introduction}

\section{Background}
\label{sec:background}
\input{sections/02-background}

\section{Data Collection}
\label{sec:data}
\input{sections/03-data}

\section{Claim Support}
\label{sec:claim-support}
\input{sections/04-claim-support}

\section{Evidence Retrieval}
\label{sec:evidence-retrieval}
\input{sections/05-evidence-retrieval}

\section{Conclusion}
\label{sec:conclusion}
\input{sections/06-conclusion}

\section*{Limitations}
\label{sec:limitations}
\input{sections/07-limitations}

\section*{Ethics}
\label{sec:ethics}
\input{sections/08-ethics}

\section*{Acknowledgments}
\label{sec:acknowledgments}
\input{sections/09-acknowledgments}

\bibliography{anthology-1,custom}
\bibliographystyle{acl_natbib}

\clearpage
\appendix

\section{Data Collection}
\label{app:annotation}
\input{appendices/annotation}

\section{Experimental Details}
\label{app:experimental-details}
\input{appendices/experimental-details}

\section{Additional Results}
\label{app:additional-results}
\input{appendices/additional-results}

\section{WiCE Comparison}
\label{app:related-work}
\input{appendices/related-work}

\end{document}

%% file: sections/00-abstract.tex
Wikipedia is a critical resource for modern NLP, serving as a rich repository of up-to-date and citation-backed information on a wide variety of subjects. The reliability of Wikipedia---its groundedness in its cited sources---is vital to this purpose. This work analyzes both how grounded Wikipedia \emph{is} and how readily fine-grained grounding evidence can be retrieved. To this end, we introduce \dataset\footnote{\url{https://github.com/wgantt/people-profiles}}---a large-scale, multi-level dataset of claim support annotations on biographical Wikipedia articles. We show that: (1) ${\sim}22\%$ of claims in Wikipedia \emph{lead} sections are unsupported by the article body; (2) ${\sim}30\%$ of claims in the article \emph{body} are unsupported by their publicly accessible sources; and (3) real-world Wikipedia citation practices often differ from documented standards. Finally, we show that complex evidence retrieval remains a challenge---even for recent reasoning rerankers.


%% file: sections/01-introduction.tex
Long an essential ingredient for LLM pretraining, Wikipedia is now widely used during inference as a repository of high-quality, citation-backed information for RAG applications \citep[][\emph{i.a.}]{lewis-etal-2020-retrieval, chen-etal-2020-bridging, fan-etal-2024-survey}. In parallel, Wikipedia has played a major role in advancing automated \emph{fact} or \emph{claim verification} \citep{dmonte-etal-2024-claim}, enabling the creation of many notable benchmarks for these tasks, such as FEVER \citep{thorne-etal-2018-fever, thorne-etal-2018-fact}, WikiFactCheck-English \citep{sathe-etal-2020-automated}, VitaminC \citep{schuster-etal-2021-get}, and WiCE \citep{kamoi-etal-2023-wice}. But whereas these works treat Wikipedia articles as sets of claims or passages to sample from for dataset curation, this work studies Wikipedia articles as \emph{whole, structured documents}---relied upon as trustworthy sources for information-seeking tasks.

First, we ask how \emph{grounded} claims in Wikipedia are. Acknowledging Wikipedia's distinction between an article's \emph{lead} section (i.e.\ the intro) and its \emph{body} (the remaining sections), we are the first to jointly explore both how claims in the lead are grounded in the body (\emph{article-internal} support; \autoref{fig:fig1}, top arrow) and how claims in the body are in turn grounded in publicly accessible cited sources (\emph{article-external} support; \autoref{fig:fig1}, bottom arrow). Second, we ask how well both standard first-stage retrievers and new reasoning-based reranking models can recover evidence about these claims---either from the body (for lead claims) or from cited source documents (for body claims). In answering these questions, we make the following contributions:

\begin{figure}
    \centering
    \includegraphics[width=\columnwidth]{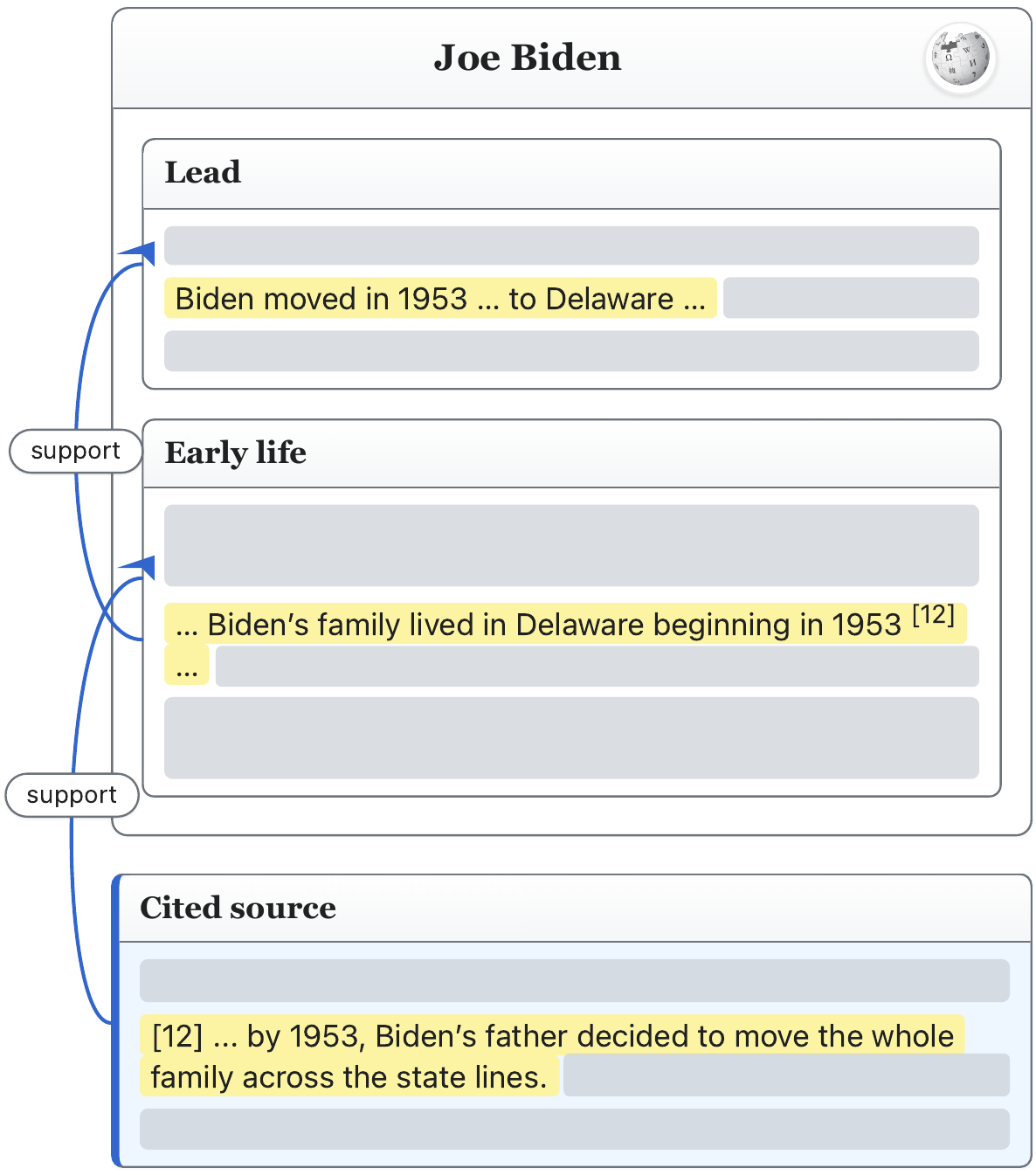}
    \caption{\dataset features fine-grained, multi-level evidential relations and scalar support labels (not shown) on Wikipedia articles---from cited sources to claims in the article body (bottom arrow), and from the body to claims in the article lead (top arrow).}
    \label{fig:fig1}
\end{figure}

\begin{itemize}
\setlength\itemsep{-0.2em}
\item We release \dataset, a new dataset of \emph{structured} Wikipedia claim support judgments for \emph{all} lead claims and \emph{all} body claims with scrapable citations from 1.5K articles about people, covering nearly 50K lead claims and 100K body claims with fine-grained scalar support labels and associated evidence.
\item We show that: (1) a surprising proportion of lead claims (${\sim}22\%$) are \emph{unsupported by the body of the same article}; (2) an even higher proportion of body claims (${\sim}30\%$) are \emph{unsupported by scrapable sources}; and, more generally, (3) actual Wikipedia citation practices often differ from documented standards.
\item We show that evidence for these claims is often \emph{complex}, involving multiple premises, and that retrieval of such evidence remains challenging even for new reasoning rerankers.
\end{itemize}

\begin{figure*}
    \centering
    \includegraphics[width=\textwidth]{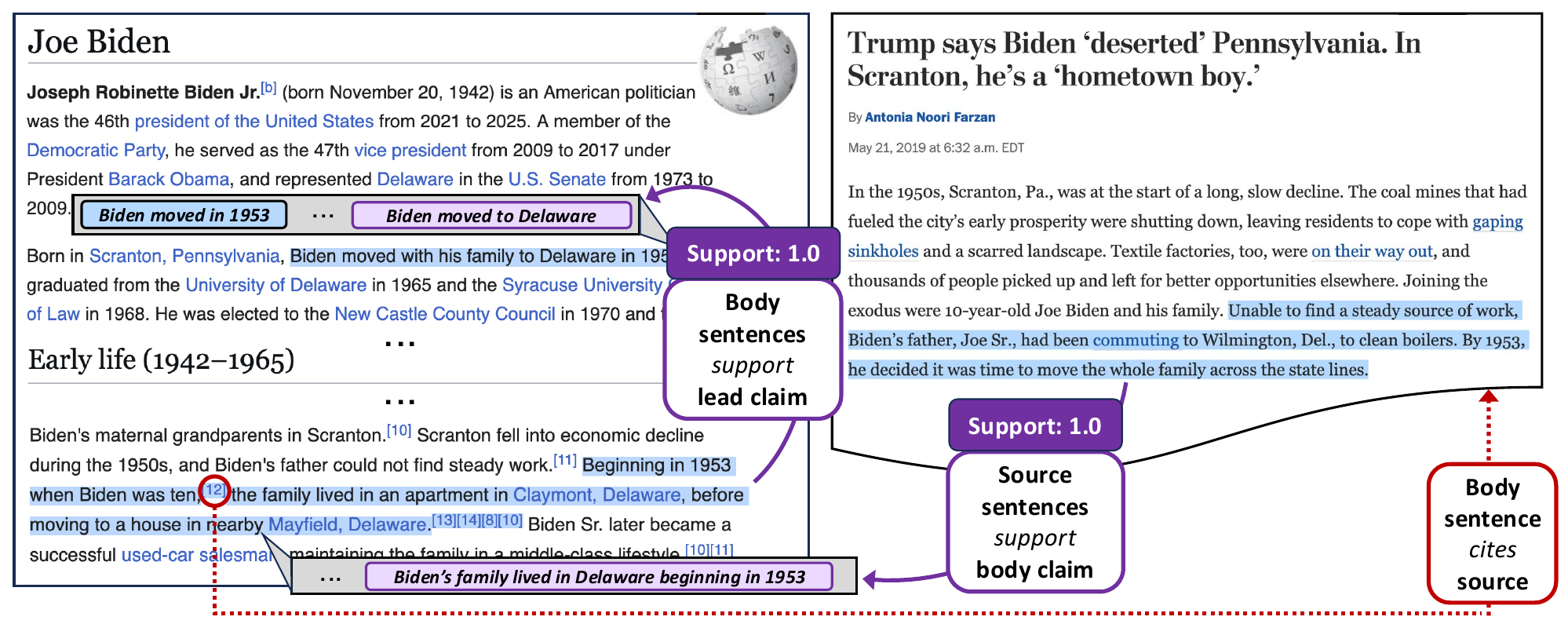}
    \caption{A more detailed view of the \emph{multi-level structure} of \dataset annotations. Claims in the \emph{lead} of a Wikipedia article (top left) are supported by sentences in the \emph{body} (bottom left), whose claims in turn are supported by evidence in cited sources (right). Prior work on Wikipedia claim verification has not considered this structure.}
    \label{fig:example}
\end{figure*}


%% file: sections/02-background.tex
\paragraph{Wikipedia-Based Claim Verification} \emph{Fact} or \emph{claim verification} is an active area of research within NLP with many supporting datasets, including several that draw heavily on Wikipedia.\footnote{We refer the reader to \citet{dmonte-etal-2024-claim} for a general survey of claim verification and focus only on Wikipedia here.} Of these, FEVER \citep[\fever;][]{thorne-etal-2018-fever, thorne-etal-2018-fact} is the most widely studied, featuring 185K claims written by human annotators, given randomly chosen Wikipedia sentences as prompts. For each claim, a different set of annotators then curated evidence passages---sets of sentences from one or more Wikipedia pages---as well as claim support labels (\textsc{Supported}, \textsc{Refuted}, \textsc{NotEnoughInfo}).

WikiFactCheck-English \citep[\wikifactcheck;][]{sathe-etal-2020-automated} consists of 125K claims also derived from Wikipedia, but uses binary (\textsc{Supported}, \textsc{Refuted}) labels. \textsc{Supported} claims are Wikipedia sentences with at least one citation and \textsc{Refuted} claims are obtained via manual edits to the original (true) claims to make them false. Evidence consists of full documents cited by the claim sentence.

VitaminC \citep[\vitaminc;][]{schuster-etal-2021-get} uses factual revisions made by Wikipedia editors on a large set of articles as well as synthetic revisions to evidence sentences from \fever to build a set of \emph{contrastive} evidence sentence pairs, based on which annotators then write \textsc{Supported} or \textsc{Refuted} claims---yielding 326K claim-evidence pairs.

Finally, the work most similar to ours is WiCE \citep[\wice;][]{kamoi-etal-2023-wice}, which annotates ternary support labels on \emph{subclaims} automatically decomposed from Wikipedia sentences drawn from the SIDE dataset \citep{petroni-etal-2023-improving}.\footnote{\autoref{tab:wice_comparison} summarizes key differences between our work and \wice. Differences with \fever/\wikifactcheck/\vitaminc are discussed in \S\ref{sec:background}.} Each subclaim is paired with annotated sets of evidence sentences from associated cited web articles. In total, WiCE contains roughly 5.4K subclaim-evidence pairs.

\paragraph{Claim Decomposition} A wealth of recent work on \emph{claim decomposition} has argued that the appropriate units for assessment of evidential support are \emph{subclaims}, i.e., sub-sentence-level propositions \citep[][\emph{i.a.}]{kamoi-etal-2023-wice, min-etal-2023-factscore, wanner-etal-2024-closer, wanner-etal-2024-dndscore, gunjal-durrett-2024-molecular}. Broadly, this argument contends that because subclaims assert single atomic propositions, they are easier and less ambiguous to evaluate than sentences, which may assert (or presuppose) multiple propositions. In practice, the optimal conception of atomicity is contested, and a variety of decomposition techniques have been proposed \citep{wanner-etal-2024-dndscore, gunjal-durrett-2024-molecular}. While we do not contribute to this debate, we accept the consensus that undergirds it---that subclaims are to be preferred to sentences---leveraging an existing claim decomposition method to construct \dataset (\S\ref{sec:data}).

\paragraph{Our Work}
Our \dataset dataset differs from the resources discussed above in several key ways. First, our claims are \emph{ecologically valid}, as they are sourced from Wikipedia articles, and not synthetically constructed or perturbed by crowdworkers (\emph{contra} \vitaminc, \wikifactcheck). Second, we annotate support on atomic \emph{subclaims}, not full sentences, making it unambiguous which proposition's support is being assessed (\emph{contra} \fever, \vitaminc, \wikifactcheck). Third, we obtain \emph{scalar} (not categorical) support judgments to capture variation in the degree of partial support that subclaims may have. Finally, our annotations capture \emph{multiple levels of evidence}---enabling us to trace support from key claims in Wikipedia lead sections down into the body, and from there into cited sources. These last two features are unique to \dataset.

%% file: sections/03-data.tex
\subsection{Methodology}
We obtain evidence for Wikipedia claims and scalar [-1,1] judgments of the degree of support/refutation for those claims given that evidence.\footnote{While refutation is uncommon in Wikipedia, we wanted to be able to capture the cases where it occurs.} We divide annotation into two phases---one for claims appearing in the article's \emph{lead} and a second for claims appearing in its \emph{body}. This is motivated by the different guidelines Wikipedia establishes for these two parts of an article: while citations are \emph{required} for contentful claims in the body (e.g.\ quotations, statistics),\footnote{See \href{https://en.wikipedia.org/wiki/Wikipedia:When_to_cite}{Wikipedia:When\_to\_cite}} ``it is common for citations to appear in the body and not the lead,'' since ``significant information should not appear in the lead if it is not covered in the remainder of the article.''\footnote{See \href{https://en.wikipedia.org/wiki/Wikipedia:Manual_of_Style/Lead_section}{Wikipedia:Manual\_of\_Style/Lead\_section}} Thus, for lead claims, we seek evidence in the body, and for body claims, we seek evidence in cited sources. Following prior work \citep{kamoi-etal-2023-wice}, we define the evidence for a claim as a set of (possibly non-contiguous) sentences. We annotate up to three sentences that together provide the strongest evidence for or against each target claim.

\subsection{Claims}
Per \S\ref{sec:background}, we adopt the view advocated in work on \emph{claim decomposition} that the appropriate units for assessment of evidential support are \emph{subclaims}, i.e., sub-sentence-level statements expressing an atomic proposition \citep[][\emph{i.a.}]{kamoi-etal-2023-wice, min-etal-2023-factscore, wanner-etal-2024-closer, wanner-etal-2024-dndscore, gunjal-durrett-2024-molecular}.\footnote{When we refer to \emph{claims} in this work, we mean \emph{subclaims} in this sense. (The claim decomposition literature often uses \emph{claim} synonymously with \emph{sentence}.)} We use the ``\textsc{DnD}'' method of \citet{wanner-etal-2024-dndscore} to jointly decompose each Wikipedia sentence into two sets of subclaims: a \emph{contextualized} set decomposed from the sentence alone and a \emph{decontextualized} set that inserts into each subclaim relevant extra-sentential context that may help resolve ambiguities in the contextualized version (e.g.\ unresolved pronouns). Annotators annotate \emph{all} subclaims decomposed from a target sentence and can toggle between the contextualized and decontextualized versions of a subclaim when identifying evidence and assessing support. Following \citet{wanner-etal-2024-dndscore}, we perform the decomposition using GPT-4o-mini \citep{gpt-4o-mini}.\footnote{See \autoref{app:annotation} for prompts and details of \textsc{DnD}.}



\setlength{\aboverulesep}{0pt}
\setlength{\belowrulesep}{0pt}
\renewcommand{\arraystretch}{1.3}
\begin{table}
    \centering
    \begin{tabular}{lccc}
        \toprule
        \rowcolor{gray!20}
        & \bf Train & \bf Dev & \bf Test \\
        \midrule
        Articles & 965 & 256 & 264 \\
        Lead Claims & 30,331 & 9,272 & 9,351 \\
        Body Claims & 60,107 & 19,712 & 18,712 \\
        Sources & 10,539 & 3,298 & 3,485 \\
        \bottomrule
    \end{tabular}
    \caption{\dataset summary statistics.}
    \label{tab:summary-statistics}
\end{table}

\subsection{Data Source}
\label{subsec:data-source}
Much prior work on claim decomposition and verification has focused on biographies of notable people---\emph{entities}---as a natural test domain in which the ground truth (e.g.\ birth and death dates, educational background) is generally uncontroversial and where sources of evidence are numerous \citep{min-etal-2023-factscore, jiang-etal-2024-core, gunjal-durrett-2024-molecular}. Given this, biographical Wikipedia pages also likely provide a rough upper bound on the degree of evidential support for claims in Wikipedia articles from other domains, which may be subject to wider debate or disagreement (e.g.\ theoretical frameworks, religions, or political movements). This upper bound would thus strengthen findings of a \emph{lack} of evidential support in these articles (\S\ref{sec:claim-support}).

Accordingly, we follow this line of prior work and annotate the Wikipedia pages for all 1,485 entities studied by \citet{min-etal-2023-factscore} and \citet{jiang-etal-2024-core}, representing a wide range of nationalities and degrees of fame. We obtain the English articles for each entity from the MegaWika 2 dataset \citep{barham-etal-2025-megawika2}, which includes (\emph{inter alia}) section boundaries, in-text citations, and citations' scraped source texts for each article. We annotate sets of evidence sentences and scalar support labels for subclaims decomposed from (1) \emph{all} sentences in articles' leads and (2) \emph{all} body sentences that bear citations to \emph{publicly accessible} sources (whose contents are provided in MegaWika 2). We focus on \emph{public} sources because neither we nor Wikipedia users---nor even RAG-enabled search engines---have the resources or licenses required to verify paywalled or print sources at scale.

\citet{barham-etal-2025-megawika2} acknowledge rare errors in their source scraping process for MegaWika 2, resulting in 404 responses or other incomplete rendering of source content. To mitigate this, we further filter out low-quality sources (e.g.\ error pages) using the DeBERTa-based \citep{he-etal-2020-deberta} text quality classifier from NVIDIA's NeMo Curator.\footnote{\url{https://github.com/NVIDIA/NeMo-Curator}}

Finally, we construct train (965 entities), dev (256), and test (264) splits via stratified sampling of entities based on their number of lead subclaims.

\subsection{Pilot Annotation}
\label{subsec:pilot-annotation}
Given the large scale of our (main) bulk annotation (${\sim}150$K claims; \S\ref{subsec:bulk-annotation}), we adopt an automatic annotation process. To verify the quality of this process, we first conduct a pilot human annotation on a collection of 160 body claims obtained from 10 entities, divided into 3 batches. Three authors participated in the pilot, with a unique pair of annotators doubly annotating each batch using an interface and instructions that we developed.\footnote{See \autoref{app:annotation} for interface details, annotation instructions, and annotator demographics.}

We then use these annotations to guide manual prompt engineering for the bulk annotation, assessing GPT-4o-mini on the same examples and aiming to optimize agreement with the human annotations along two dimensions: (1) average pairwise $\text{F}_1$ with annotated evidence sets (using exact sentence match as the underlying similarity); and (2) Krippendorff's $\alpha$ \citep[with interval difference function;][]{krippendorff-2018-content} on the scalar support labels. Model annotations from our final (best) prompt on the pilot claims yield average pairwise $\alpha{=}66.3$ and $\text{F}_1{=}62.7$ w.r.t.\ human annotations. This label agreement is only slightly below \emph{human expert} agreement reported for related works, including \fever ($\alpha{=}68.4$) and \vitaminc ($\alpha{=}70.7$), despite our use of \emph{scalar} labels and allowance of deviation in the underlying evidence sets.\footnote{These other works report agreement as Fleiss's $\kappa$, which is a special case of $\alpha$.} Agreement on evidence selection is also on par with human expert agreement on these same examples ($\text{F}_1{=}61.4$).



\subsection{Bulk Annotation}
\label{subsec:bulk-annotation}
Using GPT-4o-mini with the best prompt identified in the pilot (same for lead and body claims), we collect support and evidence annotations on all 1,485 entities. \autoref{tab:summary-statistics} shows statistics of the resulting \dataset dataset. \autoref{tab:wice_comparison} further illustrates that our automatic pipeline enables annotation of roughly 20x more body claims than \wice \citep{kamoi-etal-2023-wice}, the most similar prior work.

\begin{figure}
    \centering
    \includegraphics[width=0.85\linewidth]{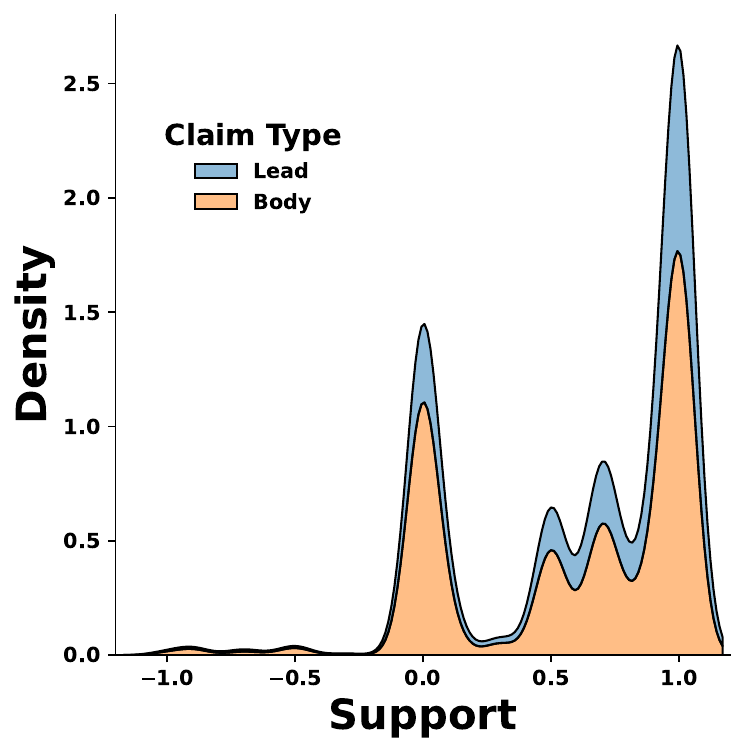}
    \vspace{-4mm}
    \caption{Kernel density estimation plots for Wikipedia lead/body claim support in the \dataset dev split. We find that many claims are \emph{not} fully grounded.\vspace{-5mm}}
    \label{fig:support-dists}
\end{figure}

%% file: sections/04-claim-support.tex
\begin{table*}
\small
    \centering
    \begin{tabular}{lll}
    \toprule
    \rowcolor{gray!20}
    \textbf{Claim Type} & \textbf{Category} & \textbf{Example Claim} \\
    \midrule
        \multirow{3}{*}{\btol} & Birth/death date/location & \emph{Josh Mansour was born in 1990.} \\
         & Nicknames & \emph{Michael Barakan is known as Shane Fontayne.} \\
         & Nationality & \emph{David Thomas Broughton is English.} \\
         & Career Status & \emph{Richie Dorman is retired.} \\
         \midrule
         \multirow{3}{*}{\stob} & Background/Summary Claims & \emph{Yorktown proved to be the last campaign of the Revolutionary War.} \\
         & Hearsay/Disputed Information & \emph{Some accounts report that Washington opposed flogging.} \\
         & Bleached Claims & \emph{There was a cabinet.} \\
    \bottomrule
    \end{tabular}
    \caption{Some common categories of lead and body claims among those with an annotated support score $\leq 0$.}
    \label{tab:unsupported-examples}
\end{table*}

In this section, we present some analysis of Wikipedia claim support enabled by \dataset, including the degree of support for both body and lead claims, the kinds of claims that tend to be unsupported, and propagation of support from body to lead.

\subsection{Support Score Distributions}
We find that \emph{significant fractions of lead and body claims are unsupported.} \autoref{fig:support-dists} plots (kernel density estimates of) lead and body claim support distributions for the \dataset dev split. We observe bimodality in both distributions, with high density around both full support (1.0) and no support (0.0). Indeed, 21.7\% of lead claims are judged \emph{unsupported} ($\leq 0$) by the body text and  29.5\% of body claims by their cited source text(s). Despite the bimodality, a sizable proportion of claims do receive partial support (i.e.\ scores in (0,1)), including 32.6\% of lead claims and 35.7\% of body claims.

\subsection{Unsupported Claims}
It is natural to wonder what sorts of claims tend to receive support scores $\leq 0$. We find that the answer differs between lead and body claims. \autoref{tab:unsupported-examples} presents several common categories of unsupported claims separately for the lead and body.

For lead claims, we observe that certain key facts---ubiquitous in lead sections---often do not receive any treatment in the body, including birth and death date and place (1st row), nicknames (2nd row), nationality (3rd row), and information about career status (4th row). In some cases (e.g.\ nicknames, nationality), the evidence itself may be difficult to obtain or arguably may not merit a citation. But in other cases (birth/death details, career status), the observed lack of body evidence seems to represent a departure from Wikipedia guidelines.\footnote{E.g.\ ``Birth and death places, if known, should be mentioned in the body of the article'' (\href{https://tinyurl.com/jkacxk85}{Wikipedia Manual of Style}).}

Categories of unsupported body claims are more varied, as article bodies themselves are more diverse. Here, some unsupported claims are ones that situate other facts that are more central to the article; others summarize (the import of) those facts (5th row). When the role of such claims is largely to contextualize or to recap ground that has been covered, the absence of supporting source material can perhaps be justified, though not always.

\emph{Hearsay} or claims implying that some fact is disputed are another important case (6th row). In the example shown, although technically only one account of Washington's opposition to flogging is required to establish that \emph{some accounts report that Washington opposed flogging}, other accounts in the source for this claim say the opposite. Cases like this---where evidence is conflicting---can thus result in support labels of 0 or less.

Finally, claim decomposition (even after decontextualization) can occasionally result in what \citet{jiang-etal-2024-core} term \emph{bleached claims}---ones that are highly likely to be true, even independent of any evidence (7th row). For this reason, bleached claims are generally not ones that require evidence.


\begin{figure}
    \centering
    \includegraphics[width=\columnwidth]{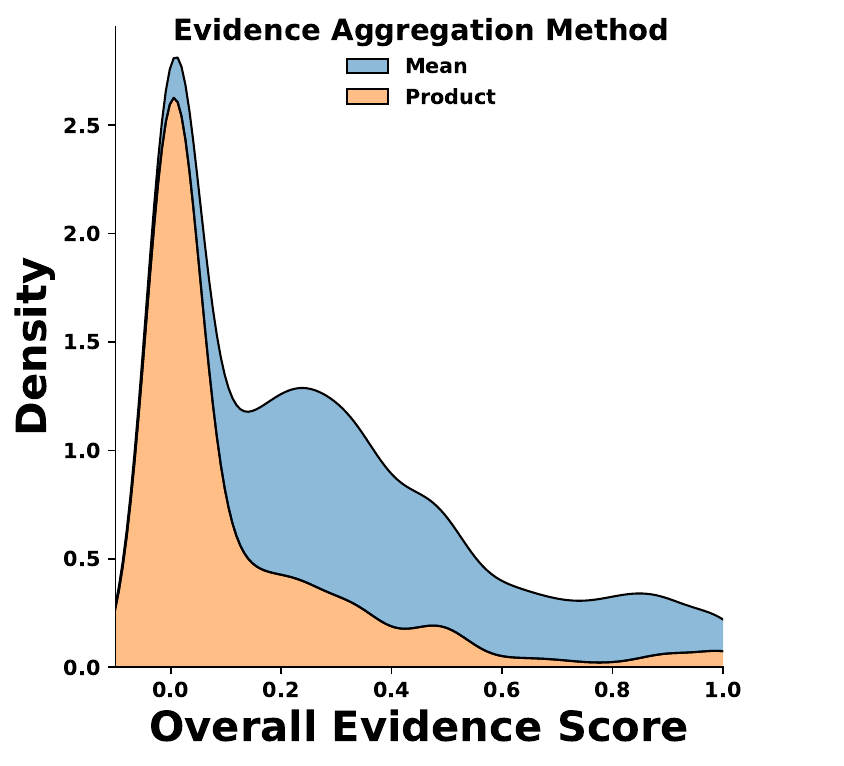}
    \caption{Distribution of overall evidence scores for \dataset dev split body evidence with mean- (blue) and product-based (orange) aggregation of body claim support scores for each evidence sentence.}
    \label{fig:support-propagation}
\end{figure}

\subsection{Claim Support Propagation}
\label{subsec:claim-support-propagation}
We annotate lead claim support \emph{given} body evidence, but we can also consider how strong the support is \emph{for that evidence} based on cited sources. We consider two methods of computing a support score for a body evidence sentence based on the support scores of the claims decomposed from it: taking the \emph{mean} of those scores or taking their \emph{product}.\footnote{For product scores, we clip body claim scores $<0$ to $0$.} We can then compute an overall score for the \emph{set} of evidence sentences for a given lead claim by applying the same aggregations (mean or product) across sentences.

First, we find that the majority of lead claims (82\%) cannot be grounded in external source material via this method---i.e.\ by attempting to locate that material via its body evidence sentences. Importantly, this does \emph{not} imply that evidence for all these lead claims is \emph{missing} from those sources. Rather, the principal issue is that, for many lead claims, many of the body sentences in the evidence set for that claim do not bear a citation to a (publicly accessible) source. Absent a citation, we thus could not annotate claims in these body sentences for source support.

This finding underscores a key challenge for \emph{all} NLP work focused on evidence attribution: real-world citation practices routinely deviate from the standard model (adopted here and elsewhere) in which \emph{each} citation-worthy sentence is expected to bear supporting citations. In practice, a citation may be provided just once at the \emph{beginning} of a paragraph (if it is clear that the whole paragraph concerns the same source) or lumped together with other citations only at the \emph{end} of a paragraph. While such practices can obscure the mapping between claims and source material, and while they may diverge from documented standards for attribution (whether in Wikipedia or elsewhere), they are nonetheless commonplace and methods should be developed to better accommodate them.

Restricting ourselves to \dataset lead claims that \emph{can} be grounded in sources, \autoref{fig:support-propagation} plots the distribution of overall evidence scores for lead claims from the dev split under both the mean (blue) and product (orange) aggregation strategies. We find very modest overall scores in both cases (an average of 0.41 for mean and 0.35 for product).

%% file: sections/05-evidence-retrieval.tex
We now turn our focus from the extent of evidential support for a target claim to retrieval of find-grained evidence for such a claim. We consider two claim-centric evidence retrieval tasks:
\begin{enumerate}[itemsep=-1ex]
    \item \btol: Retrieve all body evidence sentences for a given \emph{lead} claim
    \item \stob: Retrieve all evidence sentences from a single cited source for a given \emph{body} claim
\end{enumerate}
We also consider a third, \emph{entity-level} retrieval task:
\begin{enumerate}[itemsep=-1ex]
\setcounter{enumi}{2}
    \item \stoe: Retrieve \emph{all} evidence sentences annotated in \emph{all} cited sources for a given \emph{entity}
\end{enumerate}
We treat (1) and (2) as binary relevance tasks, aiming to recover the gold-annotated evidence sentences using the decontextualized claim as the query. For (3), we adopt fine-grained relevance labels, as different source material may be differently important to an entity's biography. Source sentences that support \emph{more} claims and support them \emph{more strongly} are assigned higher relevance.\footnote{Relevance label calculation details are in \autoref{app:experimental-details}.} For this task, we use the same query template for all entities: \emph{Tell me about the life of $\langle$entity$\rangle$, including early life, education, career, and death.}

We report first-stage retrieval results on the \dataset test set with several widely used models: BM25 \citep[sparse, lexical;][]{robertson-etal-1995-okapi}, ColBERTv2 \citep[dense, multi-vector;][]{khattab-matei-2020-colbert, santhanam-etal-2022-colbertv2}, and Stella-v5-1.5B \citep[dense, single-vector;][]{zhang-etal-2024-jasper}. We additionally report reranking results on BM25 outputs with four recent rerankers, discussed later in this section. For \btol and \stob, we report recall@\{5,10\} and NDCG@5. For \stoe, we instead report recall@100, as well as NDCG@\{5,100\}, since there are $\gg$ 3 evidence sentences per query.

\setlength{\aboverulesep}{0pt}
\setlength{\belowrulesep}{0pt}
\renewcommand{\arraystretch}{1.3}
\begin{table}
    \centering
    \small
    \begin{tabular}{lllll}
    \rowcolor{gray!20}
    \toprule
        \bf Task & \bf Model & \bf N@5 & \bf R@5 & \bf R@10 \\
    \midrule
        \multirow{3}{*}{\btol} & ColBERTv2 & 52.59 & 57.90 & 68.18 \\
        & Stella-1.5B-v5 & 59.67\textsuperscript{$\dagger$} & 64.92\textsuperscript{$\dagger$} & 75.12\textsuperscript{$\dagger$} \\
        & BM25 & 49.90 & 55.98 & 65.89 \\
        \rowcolor{blue!20}
        & RankZephyr-7B & 62.56 & 62.67 & 65.89 \\
        \rowcolor{blue!20}
        & Rank1-7B & 60.55 & 63.25 & 65.89 \\
        \rowcolor{blue!20}
        & Rank-K-7B & 63.33 & 63.84 & 65.89 \\
        \rowcolor{blue!20}
        & ReasonRank-7B & \bf 64.24 & \bf 64.49 & \bf 65.89 \\
        \rowcolor{blue!20}
        & ReasonRank-32B & 64.24 & 64.41 & 65.89 \\
        \midrule
        \multirow{3}{*}{\stob} & ColBERTv2 & 70.02 & 76.37 & 87.16 \\
        & Stella-1.5B-v5 & 74.27\textsuperscript{$\dagger$} & 80.69\textsuperscript{$\dagger$} & 90.41\textsuperscript{$\dagger$} \\
        & BM25 & 61.52 & 67.87 & 78.73 \\
        \rowcolor{blue!20}
        & RankZephyr-7B & 73.07 & 74.40 & 78.73 \\
        \rowcolor{blue!20}
        & Rank1-7B & 73.02 & 76.18 & 78.73 \\
        \rowcolor{blue!20}
        & Rank-K-32B & 75.01 & 76.28 & 78.73 \\
        \rowcolor{blue!20}
        & ReasonRank-7B & 75.92 & 76.96 & 78.73 \\
        \rowcolor{blue!20}
        & ReasonRank-32B & \bf 76.16 & \bf 77.04 & \bf 78.73 \\
        \bottomrule
    \end{tabular}
    \caption{Evidence retrieval results for the \btol and \stob tasks. White rows are first-stage retrieval results (``$^\dagger$'' indicates best results). Blue rows are reranking results on the top 10 evidence sentences from BM25 (best results are \textbf{bolded}). \textbf{N}=NDCG; \textbf{R}=Recall.}
    \label{tab:lead-and-body-retrieval-results}
\end{table}


\subsection{First Stage Retrieval}
The first three rows of \autoref{tab:lead-and-body-retrieval-results} report first stage retrieval results for the \btol and \stob tasks. On both, we obtain our best results with Stella, which shows 7+ point gains over BM25 across metrics on \btol and 11+ point gains on \stob. ColBERT also achieves notable improvements over BM25, albeit not quite as large (2+ on \btol and 8+ on \stob).

The first three rows of \autoref{tab:entity-retrieval-results} show results on the \stoe task, where both Stella and ColBERT again show large improvements over BM25 (up to +10 on NDCG@100; nearly +8 on recall@100), although Stella is no longer the clear winner. ColBERT achieves best results on NDCG@5 (24.53 vs.\ 23.79), whereas Stella achieves a slight edge on NDCG@100 (40.60 vs.\ 39.65) and best results on recall@100 (64.37 vs.\ 62.52).

Collectively, these experiments offer some preliminary evidence that effective evidence retrieval requires more than mere lexical match. While first-stage NDCG scores with dense methods (ColBERT, Stella) are decent---at least for \btol and \stob---the next section considers how much these results may be improved by leveraging yet more sophisticated \emph{reasoning} methods for reranking.

\setlength{\aboverulesep}{0pt}
\setlength{\belowrulesep}{0pt}
\renewcommand{\arraystretch}{1.3}
\begin{table}
    \centering
    \small
    \begin{tabular}{llll}
    \rowcolor{gray!20}
    \toprule
        \bf Model & \bf N@5 & \bf N@100 & \bf R@100 \\
        \midrule
        ColBERTv2 & 24.53\textsuperscript{$\dagger$} & 39.65 & 62.52 \\
        Stella-1.5B-v5 & 23.79 & 40.60\textsuperscript{$\dagger$} & 64.37\textsuperscript{$\dagger$} \\
        BM25 & 14.59 & 30.08 & 56.67 \\
        \rowcolor{blue!20}
        RankZephyr-7B & 24.91 & 35.82 & 56.67 \\
        \rowcolor{blue!20}
        Rank1-7B & 20.54 & 34.28 & 56.67 \\
        \rowcolor{blue!20}
        Rank-K-32B & \bf 26.98 & \bf 37.97 & \bf 56.67 \\
        \rowcolor{blue!20}
        ReasonRank-7B & 24.55 & 36.83 & 56.67 \\
        \rowcolor{blue!20}
        ReasonRank-32B & 26.35 & 37.60 & 56.67 \\
        \bottomrule
    \end{tabular}
    \caption{Evidence retrieval results for the \stoe task. Blue rows are reranking results on the top 100 evidence sentences from BM25. See \autoref{tab:lead-and-body-retrieval-results} for further details.}
    \label{tab:entity-retrieval-results}
\end{table}

\subsection{Reranking via Reasoning}
Recent work has shown that \emph{reasoning} rerankers, which use reasoning chains to produce relevance judgments, achieve substantial gains on other complex retrieval tasks \citep[][\emph{i.a.}]{weller-etal-2025-rank1, shao-etal-2025-reasonir, zhuang-etal-2025-rank}. Our tasks---particularly the claim-centric ones (\btol and \stob)---clearly belong in this category, as evidence may be multi-premise, with premises dispersed across the body of the article or source document. Accordingly, we evaluate several such rerankers on all three tasks:
\begin{enumerate}
\item \textbf{Rank1} \citep[7B;][]{weller-etal-2025-rank1}: A pointwise reranker based on Qwen 2.5 7B \citep{yang-etal-2025-qwen} distilled from 635K reasoning traces for MS MARCO \citep{nguyen2016ms} relevance judgments produced by DeepSeek R1 \citep{guo-etal-2025-deepseek}.
\item \textbf{Rank-K} \citep[32B;][]{yang2025rank}: A listwise reranker based on QwQ 32B \citep{team2025qwq} distilled from 50K reasoning traces from R1 on MS MARCO.
\item \textbf{ReasonRank} \citep[7B, 32B;][]{liu2025reasonrank}: Listwise rerankers based on Qwen 2.5 instruct models \citep{yang-etal-2025-qwen} distilled from R1 reasoning traces on relevance judgments across diverse domains (web search, complex QA, coding, math).
\end{enumerate}

\noindent We also evaluate \textbf{RankZephyr} \citep[7B;][]{pradeep2023rankzephyr} as a \emph{non-reasoning}, listwise LLM reranker baseline. RankZephyr is based on Zephyr\textsubscript{$\beta$} \citep{tunstall2023zephyr} and is distilled from GPT-3.5 rank lists on MS MARCO. For all models, we rerank the top 10 evidence sentences from BM25 for \btol and \stob and the top 100 for \stoe.\footnote{Thus, R@10 is unchanged between BM25 and reranking results for \btol and \stob, as is R@100 for \stoe. Additional reranking results on top of Stella rank lists are in \autoref{app:additional-results}.}

Results on \btol and \stob are shown in the blue rows of  \autoref{tab:lead-and-body-retrieval-results}, where we find large gains in NDCG@5 (10+ points) and recall@5 (6+ points) over BM25 on both tasks across all rerankers. Notably, \emph{we achieve our highest scores on} \btol \emph{and} \stob \emph{with the listwise reasoning rerankers}---Rank-K and ReasonRank---with the latter achieving best results. Interestingly, however, we do not find substantial improvements in moving from the 7B ReasonRank model to the 32B one.

Broadly, these results echo previous findings that demonstrate superior performance from listwise reasoning rerankers relative to pointwise (Rank1) and non-reasoning models (RankZephyr) on other reasoning-intensive tasks \citep[][]{yang2025rank, liu2025reasonrank}. Further, \emph{contra} \citet{weller-etal-2025-rank1}, Rank1 generally does not improve upon RankZephyr for our tasks, suggesting that \emph{both} using listwise reranking \emph{and} leveraging reasoning are key to strong performance.

\subsection{Reasoning Rerankers and Evidence Complexity}
A tempting hypothesis is that the effectiveness of listwise reasoning rerankers derives from their comparative advantage in handling more complex evidence---here, examples with multi-premise evidence sets. We explore this in \autoref{tab:retrieval-by-num-snippets}, which reports results on \btol broken down by the size of the gold evidence set (\# premises), including first-stage BM25 results and reranking results with ReasonRank-7B, Rank1 (as a \emph{pointwise} baseline), and RankZephyr (as a \emph{non-reasoning} baseline).

The story these results tell is more complicated than the hypothesis suggests. Although we do find that ReasonRank achieves larger gains over BM25 for 2-premise (+18.3 N@5, +12.4 R@5) and 3-premise (+12.8 N@5, +9.1 R@5) evidence sets than for 1-premise sets (+12.6 N@5, +4.8 R@5), the same is also true of RankZephyr (which does not use reasoning) and Rank1 (which is not listwise). Rather, since ReasonRank achieves the best results among models over \emph{all} evidence set sizes, a better explanation is that this model simply exhibits the strongest ability to reason about evidential support \emph{full stop}, irrespective of evidence complexity.

Lastly, we note the large drops in performance observed across all models as one moves from 1-, to 2-, to 3-premise evidence sets. This suggests that there remains substantial room for better cross-premise reasoning, even among the sophisticated listwise rerankers studied here.





\subsection{Retrieval Difficulty by Task}
A final notable observation about our retrieval results is that \emph{retrieval difficulty varies widely across tasks}. We obtain highest N@5 scores on \stob, followed by \btol and then by \stoe. Intriguingly, this ranking tracks the granularity of the queries, where body claims (\stob) tend to provide the most detailed information, lead claims (\btol) present key high-level facts, and entity-level queries (\stoe) are most general, inquiring about the life of the target entity in generic terms. Intuitively, the detailed claims appearing in Wikipedia body sections tend to bear greater lexical and semantic similarity to their supporting source material than the higher-level queries of \btol or \stoe do to theirs. At least for \stob and \stoe, an initial \emph{passage}-level retrieval stage could plausibly yield improved recall, but the core challenge of reasoning over combinations of fine-grained (sentence-level) premises would still remain.



\setlength{\aboverulesep}{0pt}
\setlength{\belowrulesep}{0pt}
\renewcommand{\arraystretch}{1.3}
\begin{table}
    \centering
    \small
    \begin{tabular}{cllll}
        \toprule
        \rowcolor{gray!20}
        \bf \#Sents & \bf Model & \bf N@5 & \bf R@5 & \bf R@10 \\
        \midrule
        \bf 1 & BM25 & 76.13 & 86.60 & 91.78 \\
        \rowcolor{blue!20}
        & RankZephyr-7B & 86.53 & 89.64 & 91.78 \\
        \rowcolor{blue!20}
        & Rank1-7B & 84.49 & 90.67 & 91.78 \\
        \rowcolor{blue!20}
        & ReasonRank-7B & \bf 88.68 & \bf 91.38 & \bf 91.78 \\
        \midrule
        \bf 2 & BM25 & 47.29 & 53.18 & 66.51 \\
        \rowcolor{blue!20}
        & RankZephyr-7B & 63.90 & 63.28 & 66.51 \\
        \rowcolor{blue!20}
        & Rank1-7B & 61.32 & 63.47 & 66.51 \\
        \rowcolor{blue!20}
        & ReasonRank-7B & \bf 65.62 & \bf 65.57 & \bf 66.51 \\
        \midrule
        \bf 3 & BM25 & 25.25 & 27.26 & 39.11 \\
        \rowcolor{blue!20}
        & RankZephyr-7B & 37.01 & 34.88 & 39.11 \\
        \rowcolor{blue!20}
        & Rank1-7B & 35.40 & 35.28 & 39.11 \\
        \rowcolor{blue!20}
        & ReasonRank-7B & \bf 38.08 & \bf 36.37 & \bf 39.11 \\
        \midrule
    \end{tabular}
    \caption{Select retrieval and reranking results on \btol broken down by number of gold evidence sentences.}
    \label{tab:retrieval-by-num-snippets}
\end{table}

%% file: sections/06-conclusion.tex
We have presented a study of evidential support and retrieval on Wikipedia and have introduced \dataset, a large dataset of fine-grained, multi-level evidential support annotations on nearly 1,500 Wikipedia articles and their cited sources. We have shown that: (1) a sizable fraction of Wikipedia \emph{lead} claims are unsupported by their body sections, and Wikipedia \emph{body} claims by their (publicly accessible) cited sources; (2) evidence retrieval for these claims grows much more challenging as the generality of the queries (\btol $\rightarrow$ \stob $\rightarrow$ \stoe) and the evidence complexity (1 $\rightarrow$ 2 $\rightarrow$ 3 premises) increases; and lastly (3) new reasoning-based rerankers enable much more effective retrieval of complex evidence relative to traditional (sparse or dense) methods. We release \dataset to aid future work on claim verification and on advancing understanding of Wikipedia as a knowledge source for modern NLP.

%% file: sections/07-limitations.tex
We acknowledge several limitations of our work. First, \dataset focuses only on Wikipedia articles about people. We chose this focus because biographies tend to have a higher proportion of uncontroversial facts relative to other domains (e.g.\ concepts or events) and because multiple prior works in this area also focus on people \citep{min-etal-2023-factscore, jiang-etal-2024-core, gunjal-durrett-2024-molecular}. However, it is possible support distributions or evidence retrieval difficulty could differ in other domains. (Granted, we believe domain shift would likely \emph{strengthen} many of the claims we make, as discussed in \S\ref{subsec:data-source}.)

Second, as we emphasize in the main text, our claims about evidential support extend only to \emph{publicly accessible, digital sources}. We therefore cannot make conclusions about support \emph{across all source types} in Wikipedia. The number of paid licenses and the level of access to print resources required for this are simply infeasible to obtain.

Third, we leverage GPT-4o-mini as an annotator to facilitate our large-scale bulk data collection. While the agreement we observe between this model and our human annotations is strong (\S\ref{sec:data}), LLMs have their own response biases and may not be fully calibrated when providing scalar judgments \citep{lovering-etal-2024-language}.


Finally, as we note in \S\ref{subsec:claim-support-propagation}, real-world citation practices can differ from what is espoused in documented standards---where citations often do not appear on a sentence that would seem to require one, but where they may appear elsewhere in the same paragraph. Since we restrict our annotations to body sentences that \emph{do} bear citations, we thus cannot draw conclusions about sentences whose source may appear as a citation on some \emph{other} sentence. However, we suspect that our estimate of the proportion of supported body claims decomposed from \emph{citation-bearing} sentences is an upper bound on the proportion we would obtain for all \emph{citation-worthy} sentences.

%% file: sections/08-ethics.tex
\dataset's use of sources from MegaWika 2.0 and our release of this data (via a CC-BY-4.0-SA license) is consistent with MegaWika 2's own CC-BY-4.0-SA license. Our principle transformation of the original Wikipedia articles consists in the claim decomposition, which is performed by an LLM (GPT-4o-mini), and which can occasionally result in (sub)claims that misrepresent the article's original content and thus (potentially) facts about the subject. Although our claim decompositions are generally very faithful to the original texts, users of \dataset should be aware of this possibility.

%% file: sections/09-acknowledgments.tex
The authors would like to thank Alex Martin for help testing out the annotation interface, as well as colleagues at the Human Language Technology Center of Excellence for general comments and feedback on this work.

%% file: appendices/annotation.tex
\subsection{Annotator Demographics}
The first three authors of this work---all native English-speaking graduate students in NLP or professional NLP researchers---conducted the human pilot annotations. These authors also jointly produced the annotation instructions beforehand. None was compensated beyond their co-authorship on this work.

\subsection{Claim Decomposition}
\label{app:annotation::claim-decomposition}
Decomposition is the process of breaking down sentences into simpler, atomic components---aiming to isolate individual, independent claims for downstream applications. A common approach to claim decomposition uses LLMs to segment a sentence into independent facts, each containing one ``piece'' of information. However, these subclaims can be ambiguous, with vague references that are uninterpretable without the context of the document. The process of \emph{decontextualization} mitigates this issue by rephrasing a subclaim such that it is fully intelligible as a standalone statement, without the original document as context. These two processes are complementary: decomposition divides sentences into smaller parts, whereas decontextualization adds information. 

We use the ``DnD'' decomposition and decontextualization method introduced by \citet{wanner-etal-2024-dndscore}, which uses an LLM prompt-based method for obtaining subclaim decompositions and the corresponding decontextualized subclaims. We decompose and decontextualize sentences from the original Wikipedia page, either from the lead (in the \btol{} task) or body (in the \stob{} task), and provide the lead paragraph (\btol{}) or additionally the body paragraph from which the claim originates (\stob{}) as context for decontextualization. During the pilot annotation, annotators are able to toggle between the subclaim and its decontextualized version to select up to three sentences that together provide the \emph{strongest} evidence (either supporting or refuting) for the subclaim. Finally, after identifying evidence, annotators determine a support score for the subclaim given that evidence. The bulk annotation provides only the decontextualized subclaim in the prompt. Following \citeauthor{wanner-etal-2024-dndscore}, we use GPT-4o-mini \citep{gpt-4o-mini} to perform DnD.

\subsection{Annotation Interface}
\begin{figure*}
    \centering
    \includegraphics[width=\linewidth]{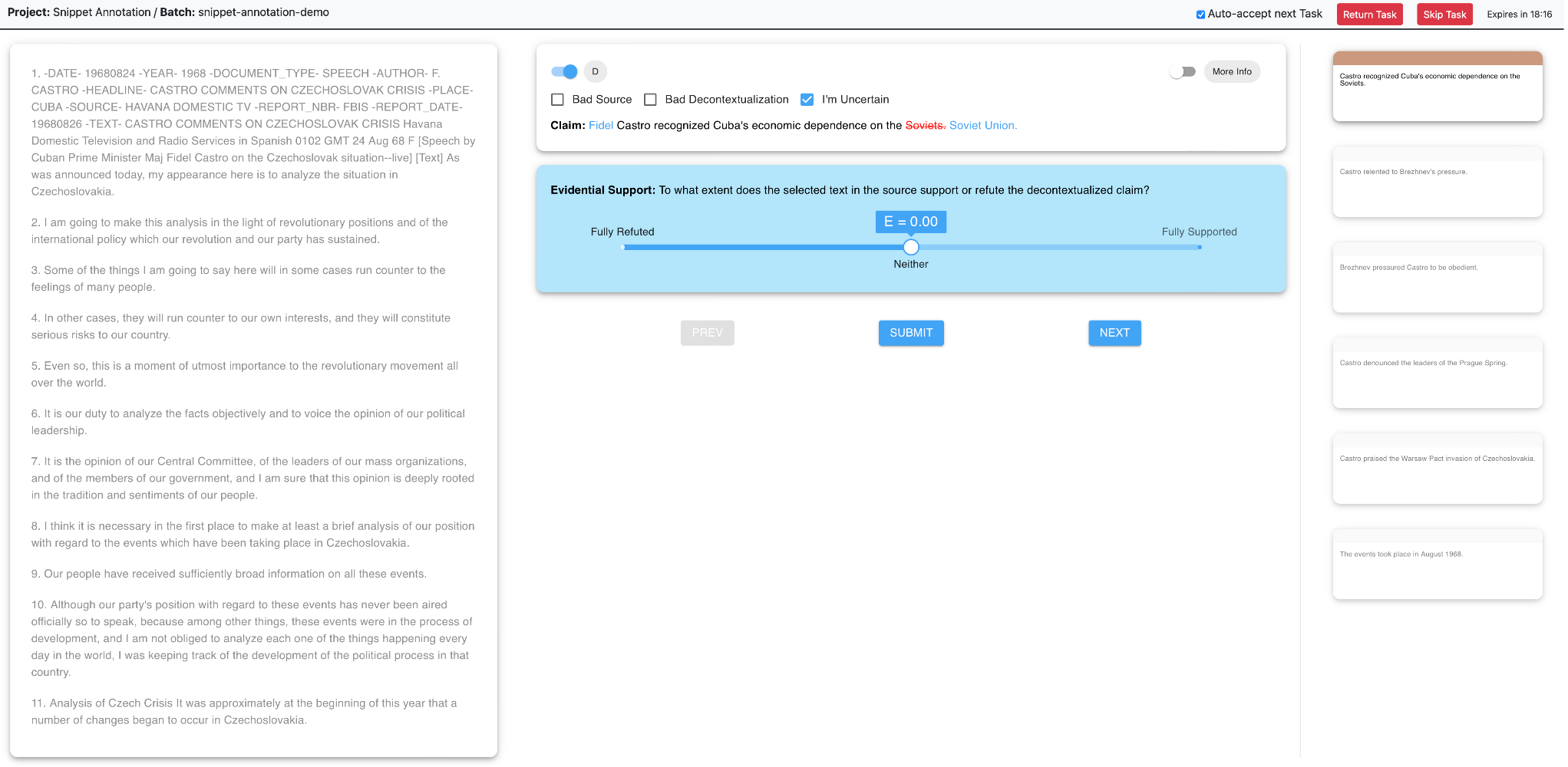}
    \caption{Annotation interface for the human pilot annotation. Detailed description can be found in Appendix \ref{app:annotation::claim-decomposition}.}
    \label{fig:annotation-interface}
\end{figure*}

The annotation interface used for the human annotation is shown in \autoref{fig:annotation-interface}. The full, sentence-split text of a cited source article is shown on the far left. All of the subclaims decomposed from a single Wikipedia body sentence citing that source article are shown in a vertical list of tiles on the far right, with the currently selected subclaim displayed in the top middle part of the screen (to the right of ``\textbf{Claim:}''). Here, annotators can toggle between the original and decontextualized versions of the subclaim using the \textbf{D} toggle shown above the subclaim, with differences (additions, deletions) between the decontextualized and original versions shown in blue and red. Annotators can also display the sentence that the current subclaim was decomposed from, along with its full Wikipedia context, by clicking the \textbf{More Info} toggle in the top right.

Several checkboxes are also shown above the subclaim to enable annotators to indicate that:
\begin{itemize}
    \item The source text is uninterpretable or otherwise low quality (\textbf{Bad Source})
    \item The subclaim is unfaithful to the meaning of the sentence from which it was decomposed (\textbf{Bad Decontextualization})
    \item It is simply too difficult to determine how the current subclaim relates to the source material---e.g.\ because the source document is too technical for the annotator to understand (\textbf{I'm Uncertain})
\end{itemize}

Annotators select up to three sentences from the source text on the left that together provide the strongest evidence (either supporting or refuting) for the target subclaim. We chose a maximum of three sentences because this enabled adequate coverage of the evidence for the vast majority of claims while keeping the task tractable for annotators.

Finally, the blue box (bottom middle) is used to specify the support score for the currently selected subclaim, given the identified evidence. After selecting evidence and providing a support score for all subclaims (toggling between them using the NEXT and BACK buttons on bottom), annotators submit their work via the SUBMIT button.

\subsection{Prompts and Hyperparameters}

The prompt used for bulk annotation with GPT-4o-mini is shown in \autoref{fig:ann_prompt_a} through \autoref{fig:ann_prompt_e} (divided over multiple pages due to the length of the instructions). This prompt was selected based on highest agreement with the human pilot annotations after numerous manual iterations on other prompts. We used \texttt{gpt-4o-mini-2024-07-18}, the most recent version of the model available at the time. 
Annotations were generated with temperature 0, with a limit of 2K output tokens to accommodate source texts of up to 126K tokens. Source texts exceeding this limit were truncated, though this was rarely required.

%% file: appendices/experimental-details.tex
\subsection{Qrels for \stoe}
For the \stoe task in \S\ref{sec:evidence-retrieval}, we assign fine-grained relevance labels to sentences in the source documents for a given entity based on (1) how \emph{strongly} they support a Wikipedia body claim, (2) how many body claims they support, (3) how strongly they support lead claims \emph{via} body claims, and (4) how many lead claims they support.

Given an article for entity $E$, a sentence $S_B$ in the article's body, a  sentence $S_S$ in some cited source, and a claim $C$, we define the following:
\begin{itemize}
    \item $lead_E(S_B)$: the set of \emph{lead} claims that have $S_B$ in their (body) evidence set
    \item $body_E(S_S)$: the set of \emph{body} claims that have $S_S$ in their (source) evidence set
    \item $support(C)$: the support score for a claim $C$
    \item $sent(C)$: the sentence that claim $C$ was decomposed from
\end{itemize}

Letting $C_B$ be a body claim and $C_L$ be a lead claim, we then define the relevance of a source sentence $S_S$ to a query $Q_E$ about entity $E$ as the following weighted sum:

\begin{eqnarray*}
    Rel(Q_E,S_S) &= \sum_{C_B \in body_E(S_S)} w_{C_B} \\& \cdot \text{abs}(support(C_B))\\
    w_{C_B} &= 1 + \sum_{C_L \in lead_E(sent(C_B))} \\& \text{abs}(support(C_L))
\end{eqnarray*}

Intuitively, $Rel(Q_E, S_S)$ is a weighted sum of the absolute values of the support scores of \emph{all} body claims ($C_B$'s) that $S$ is evidence for ($body_E(S_S)$). We use the absolute value of the support score because $S$ is equally important as evidence regardless of whether it is supporting or refuting evidence.

The weight $w_{C_B}$ associated with each body claim $C_B$ is 1, plus the sum of (absolute values of) support scores of all \emph{lead} claims for which $sent(C_B)$---the sentence $C_B$ was decomposed from---provides evidence. This rewards $S$ for \emph{indirectly} supporting a lead claim $C_L$ \emph{via a body claim} ($C_B$), proportional to the degree of support for $C_L$. The motivation here is simply that (1) lead claims typically represent more important facts about an entity than body claims, and thus sentences that (indirectly) provide evidence for them should be rewarded, and (2) that reward should be proportional to the degree of support.

We note that this is a somewhat heuristic weighting scheme, as $C_B$ is given credit merely for being \emph{decomposed from} a sentence that supports a lead claim $C_L$---even if a \emph{different} claim ($C_B'$) decomposed from the same sentence provides the bulk of the evidence for $C_L$. Collecting further annotations to enable more precise assignment of relevance scores is a direction we are pursuing for future work.

\subsection{Retrieval Model Details}

For BM25 (no parameters), we use the implementation provided in the \texttt{bm25s} library \citep{lu-2024-bm25s} with default settings. We access Stella-1.5B-v5 (1.5 billion parameters) through the \texttt{sentence-transformers} library with default settings (i.e.\ no hyperparameter search was performed). Finally we access ColBERTv2 (\texttt{jinaai/jina-colbert-v2} on HuggingFace; 559M parameters) via the \texttt{ragatouille} library\footnote{\url{https://github.com/AnswerDotAI/RAGatouille}}, leveraging FAISS for indexing \citep{johnson-etal-2019-billion}, and again using default settings. Neither Stella-1.5B-v5 nor ColBERTv2 were fine-tuned on \dataset. All experiments were carried out on a single NVIDIA A100 GPU except the reranking experiments, for which four A100s were used. All main text results reflect single runs.


Rank1 outputs were generated with temperature 0 and a maximum of 8,192 output tokens. We adopted the generation temperatures for listwise rerankers from the defaults listed in their public repositories and limited their outputs to 8K tokens. For listwise reranking, we used a window size of 20 with stride 10.

\subsection{Reranking Prompts}
\autoref{fig:prompt-rank1-a}, \autoref{fig:prompt-rank1-b}, \autoref{fig:prompt-listwise-a}, and \autoref{fig:prompt-listwise-b} list the task-specific modifications made to the prompts used in the original reranker implementations.

\subsection{Use of AI Assistants}
No AI assistance was used in the ideation or in the writing of this paper. GitHub Copilot was used to assist in writing the code for some of the experiments and analysis.

%% file: appendices/additional-results.tex
\subsection{Additional Reranking Results} \autoref{tab:lead-and-body-more-results} shows select reranking results ($k=10$) on top of Stella-1.5B-v5 outputs (rather than BM25, as in the main text) for the \btol and \stob tasks. Similar to the results reported in \autoref{tab:lead-and-body-retrieval-results}, we obtain the highest scores with ReasonRank on both tasks, with comparably large gains over first-stage retrieval.

\autoref{tab:entity-more-results} shows the same results for the \stoe task ($k=100$). Here again, ReasonRank achieves the best scores, although the gains relative to first-stage retrieval are much smaller compared to those based on BM25 outputs (\autoref{tab:entity-retrieval-results}).

\setlength{\aboverulesep}{0pt}
\setlength{\belowrulesep}{0pt}
\renewcommand{\arraystretch}{1.3}
\begin{table}
    \centering
    \small
    \begin{tabular}{lllll}
    \rowcolor{gray!20}
    \toprule
        \bf Task & \bf Model & \bf N@5 & \bf R@5 & \bf R@10 \\
    \midrule
        \btol & Stella-1.5B-v5 & 59.67 & 64.92 & 75.12 \\
        \rowcolor{blue!20}
        & RankZephyr-7B & 68.96 & 69.92 & 75.12 \\
        \rowcolor{blue!20}
        & Rank-K-32B & 70.0 & 71.44 & 75.12 \\
        \rowcolor{blue!20}
        & ReasonRank-7B & \bf 71.11 & \bf 72.36 & \bf 75.12 \\
        \midrule
        \stob & Stella-1.5B-v5 & 74.27 & 80.69 & 90.41 \\
        \rowcolor{blue!20}
        & RankZephyr-7B & 81.28 & 83.88 & 90.41 \\
        \rowcolor{blue!20}
        & Rank-K-32B & 83.54 & 86.40 & 90.41 \\
        \rowcolor{blue!20}
        & ReasonRank-7B & \bf 84.31 & \bf 87.22 & \bf 90.41 \\
        \bottomrule
    \end{tabular}
    \caption{Reranking results for select models on outputs from Stella-1.5B-v5 on the \btol and \stob tasks (best results are \textbf{bolded}). \textbf{N}=NDCG; \textbf{R}=Recall.}
    \label{tab:lead-and-body-more-results}
\end{table}

\setlength{\aboverulesep}{0pt}
\setlength{\belowrulesep}{0pt}
\renewcommand{\arraystretch}{1.3}
\begin{table}
    \centering
    \small
    \begin{tabular}{llll}
    \rowcolor{gray!20}
    \toprule
        \bf Model & \bf N@5 & \bf N@100 & \bf R@100 \\
        \midrule
        Stella-1.5B-v5 & 23.79 & 40.60 & 64.37 \\
        \rowcolor{blue!20}
        RankZephyr-7B & 25.28 & 41.14 & 64.37 \\
        \rowcolor{blue!20}
        Rank-K-32B & 25.47 & 42.08 & 64.37 \\
        \rowcolor{blue!20}
        ReasonRank-7B & \bf 25.99 & \bf 42.54 & \bf 64.37 \\
        \bottomrule
    \end{tabular}
    \caption{Reranking results for select models on outputs from Stella-1.5B-v5 on the \stoe task (best results are \textbf{bolded}). \textbf{N}=NDCG; \textbf{R}=Recall.}
    \label{tab:entity-more-results}
\end{table}

%% file: appendices/related-work.tex
\autoref{tab:wice_comparison} summarizes key differences between our \dataset and the most similar existing benchmark, WiCE \citep{kamoi-etal-2023-wice}. See \S\ref{sec:background} for a detailed comparison between \dataset, WiCE, and other related works.

\setlength{\aboverulesep}{0pt}
\setlength{\belowrulesep}{0pt}
\renewcommand{\arraystretch}{1.3}
\begin{table*}[ht]
\centering
\begin{tabularx}{0.9\linewidth}{
    >{\raggedright\arraybackslash}X 
    >{\centering\arraybackslash}m{0.8cm} 
    >{\centering\arraybackslash}m{2.2cm} 
    >{\centering\arraybackslash}m{3cm}
}
\toprule
\rowcolor{gray!20}
\textbf{Dataset Characteristic} & \textbf{Split} & \textbf{WiCE} & \textbf{\dataset{} (Ours)} \\
\midrule
Support Scores & --- & Categorical & Scalar \\
\midrule
Article-\textbf{internal} grounding annotations & --- & {\color{red}\ding{55}} & {\color{ForestGreen}\ding{51}} \\
\midrule
Article-\textbf{external} grounding annotations & --- & {\color{ForestGreen}\ding{51}} & {\color{ForestGreen}\ding{51}} \\
\midrule
Subset of article-\textbf{external} subclaims annotated & --- & SIDE subset \citep{petroni2022side} & All available \\
\midrule
\multirow{3}{=}{Annotations per subclaim} 
    & Train & 3 human & 1 LLM  \\
    & Dev & 5 human & 1 LLM  \\
    & Test & 5 human & 1 LLM \\
\midrule
\multirow{3}{=}{Number of body subclaims} 
    & Train & 3,470 & 60,107 \\
    & Dev & 949 & 19,712 \\
    & Test & 958 & 18,712 \\
\bottomrule
\end{tabularx}
\caption{Comparison of dataset characteristics between \dataset{} and WiCE, the most similar prior work. \dataset{} features fine-grained \emph{scalar} support labels, \emph{article-internal} support relations, and ${\sim}20$x more examples than WiCE.}
\label{tab:wice_comparison}
\end{table*}

\clearpage
\begin{figure*}
\begin{tcolorbox}[colback={gray!20},title={\textbf{\dataset Annotation Prompt}},colbacktitle=white,coltitle=black]
In this task, you will be shown a claim along with a list of sentences representing a document that might provide evidence for the claim. Given this information, you will perform two steps, described below. 

\vspace*{\baselineskip}
For both steps, rely on the following two definitions of evidence:

Definition 1: “Supporting evidence”:

A set of sentences S provides supporting evidence for a claim c if, supposing the contents of S were true, it would give you greater reason to believe that c is true, all else equal.

Definition 2: “Refuting evidence”:

A set of sentences S provides refuting evidence for a claim c if, supposing the contents of S were true, it would give you greater reason to believe that c is false, all else equal.

\vspace*{\baselineskip}
Step 1:

\vspace*{\baselineskip}
Select 0, 1, 2, or *at maximum* 3 sentence(s) from the document that provide the strongest supporting evidence or refuting evidence for the claim. If no sentences in the document provide evidence, do not select any sentences. 

\vspace*{\baselineskip}
Additional guidelines for Step 1:

(a) You may need to use logic and common sense to *infer* that a sentence provides evidence for the claim. For example, you can use common sense to assume that a person wearing reading glasses struggles with their sight. 

(b) Do not assume any parts of the claim are common knowledge. You must find evidence for all parts of the claim. For example, if the claim states that Vidya, the English chef, has poor vision, you would need to find evidence that Vidya is English and a chef, as well.

(c) A sentence might provide evidence for the claim only when combined with other sentences. For example, if Sentence A states Bob is married to Mary, and Sentence B states that Mary is a doctor, Sentences A and B together provide supporting evidence for the claim that Bob has a doctor in his family.

(d) Please make sure the entities and events in your selected sentences match those in the claim. For example, dates and names, as determined by the rest of the document, should match the claim; else, the sentences do not provide evidence.

\end{tcolorbox} 
\caption{}
\label{fig:ann_prompt_a}
\end{figure*}

\begin{figure*}
\begin{tcolorbox}[colback={gray!20},title={\textbf{\dataset Annotation Prompt, continued}},colbacktitle=white,coltitle=black]

Step 2:

\vspace*{\baselineskip}
Given your selected set of sentences from Step 1, score the degree to which these sentences (taken together) support or refute the claim. Determine the score according to the following definition of a scale from -1 to 1:

\vspace*{\baselineskip}
-1: The claim is *fully refuted*: The claim would have to be false, supposing the sentences you selected were true.

Scores between -1 and 0 (-0.9, -0.8, -0.7, -0.6, -0.5, -0.4, -0.3, -0.2, -0.1): The claim is *partially refuted*. The claim would have to be false, but some parts are likely true.

0: The claim is neither supported nor refuted. It is equally likely to be true or false.

Scores between 0 and 1 (0.1, 0.2, 0.3, 0.4, 0.5, 0.6, 0.7, 0.8, 0.9): The claim is *partially supported*. The claim is likely partially true, with missing evidence. No parts of the claim are likely to be false.

1: The claim is *fully supported*: The claim would have to be fully true, supposing the sentences you selected were true.

\vspace*{\baselineskip}
Additional guidelines for Step 2:

(a) Use only the content of your selected sentences to make your judgment. Do not use any knowledge you may already have about the claim, nor any context from other sentences in the document. For example, even if you know that London is in England, or it is stated elsewhere in the document, you cannot judge that detail of the claim as supported unless it is stated in your selected sentences.

(b) As in Part 1, do not assume any parts of the claim are common knowledge. Assign the score based on all parts of the claim, even if they seem obviously true or false.

(c) The document might only contain evidence for a similar but distinct claim. For example, if the strongest evidence states that the president ate at a restaurant on a Friday, this is not refuting evidence for the claim that the president ate at a restaurant on Tuesday; in fact, there is no evidence to support or refute the claim.

\vspace*{\baselineskip}

\end{tcolorbox} 
\caption{}
\label{fig:ann_prompt_b}
\end{figure*}

\begin{figure*}
\begin{tcolorbox}[colback={gray!20},title={\textbf{\dataset Annotation Prompt, continued}},colbacktitle=white,coltitle=black]
Below are 10 examples of scoring sentences that have already been selected from a document as supporting or refuting evidence for a claim:

\vspace*{\baselineskip}
\#\#\#Example 1\#\#\#

Claim: "Methane Momma is a short film directed by Alain Rimbert." 

Selected sentences: ["Well, good news \u2013 last week, in the middle of one of the worst heat waves that New York has seen in recent memory, a pajama-clad (and still ripped) Van Peebles entered ex-Sun Ra bandmember Spaceman's Harlem-based studio and recorded his last takes on the rambling poem he's entitled \"Methane Momma.\""] 

Score: -0.7

\vspace*{\baselineskip}
\#\#\#Example 2\#\#\#

Claim: "Raj Kapoor was hospitalised for about a month."

Selected sentences: ["Suddenly, Kapoor collapsed, and was rushed to the All India Institute of Medical Sciences for treatment.", 
"The country's top cardiologists tried their best, but could not save him."]

Score: -0.1

\vspace*{\baselineskip}
\#\#\#Example 3\#\#\#

Claim: "Ottawa is a city located in the province of Ontario, Canada, and is where Matthew Perry attended school."

Selected sentences: []

Score: 0

\vspace*{\baselineskip}
\#\#\#Example 4\#\#\#

Claim: "Paul Thomas Anderson registered himself with the Writers Guild of America under the name 'Paul Anderson.'"

Selected sentences: []

Score: 0

\vspace*{\baselineskip}
\#\#\#Example 5\#\#\#

Claim: "There were exile forces opposing Idi Amin's regime."

Selected sentences: ["Since leading his guerrilla forces to Kampala in 1986, his most impressive flexibility has been his capacity to present two concurrent faces: one is that of the democratic reformer, the other is of the fearsome military ruler.",
"The former is the saviour of Uganda's post-colonial collapse under presidents Milton Obote and Idi Amin, patron of democracy, and emancipator of woman and ethnic and religious minorities."]

Score: 0.1

\end{tcolorbox}
\caption{}
\label{fig:ann_prompt_c}
\end{figure*}

\begin{figure*}
\begin{tcolorbox}[colback={gray!20},title={\textbf{\dataset Annotation Prompt, continued}},colbacktitle=white,coltitle=black]
\#\#\#Example 6\#\#\#

Claim: "Margaret Rose Vendryes wrote about Richmond Barth\u00e9's work further in her 2008 book."

Selected sentences: ["By coincidence, Dr. Vendryes was the Schomburg's scholar-in-residence and was researching her Princeton doctorate thesis on Barthe, which evolved into her landmark book Casting Feral Benga: A Biography of Richmond Barth\u00e9's Signature Work."] 

Score: 0.3

\vspace*{\baselineskip}
\#\#\#Example 7\#\#\#

Claim: "Margaret Rose Vendryes gave a lecture in 2015." 

Selected sentences: ["This Thursday, February 5 at the Jepson Center, Dr. Vendryes will give the opening lecture for the exhibition."]

Score: 0.5

\vspace*{\baselineskip}
\#\#\#Example 8\#\#\#

Claim: "The exhibit presented by The New York Public Library for the Performing Arts was extensive."

Selected sentences: ["Curated by Doug Reside, the Lewis B. and Dorothy Cullman curator of the library's Billy Rose Theatre Division, the installation will run through March 31, 2020, and feature original costumes, set models, and archival video tied to Prince's productions, including models for several productions.",
"The full display will honor the more than six-decade legacy of Prince.",
"An open cabaret stage will allow viewers to perform songs from his shows or record their own stories about their experience with Prince's theatrical work to add to the live nature of the homage."]

Score: 0.7

\vspace*{\baselineskip}
\#\#\#Example 9\#\#\#

Claim: "The location of Matthew Perry's funeral was Forest Lawn Memorial Park (Hollywood Hills), a cemetery."

Selected sentences: ["Photo: David M. Benett/Dave Benett/Getty Matthew Perry's loved ones gathered for the actor's funeral on Friday.", 
"The service was held at Forest Lawn Memorial Park in Los Angeles near Warner Bros. Studios.,"]
Score: 0.9

\vspace*{\baselineskip}

\#\#\#Example 10\#\#\#

Claim: "The promotional video was 60 minutes long."

Selected sentences: ["Microsoft made a \"cyber sitcom\" to promote it.",
"The final product [debuted on VHS on August 1, 1995](https://books.google.com/books?id=0QsEAAAAMBAJ\&\\lpg=RA1-PA62\&dq=matthew\%20perry\%20jennifer\%20aniston\%20windows\%2095\&pg=RA1-PA62\#v=onepage\&q\&f=false), satisfying everybody who wished Friends were an hour long, had four fewer friends, and involved a guide to file management."]

Score: 1

\end{tcolorbox}
\caption{}
\label{fig:ann_prompt_d}
\end{figure*}

\begin{figure*}
\begin{tcolorbox}[colback={gray!20},title={\textbf{\dataset Annotation Prompt, continued}},colbacktitle=white,coltitle=black]
Finally, here are the claim and list of document sentences for your task:

\vspace*{\baselineskip}
Claim: <subclaim>

\vspace*{\baselineskip}
Document sentences:

<numbered source sentences>

\vspace*{\baselineskip}
Write your response in a dictionary in the format shown below. Write the dictionary and nothing else.

Dictionary format:

{"sentences": [

        "[<sentence number>] <sentence selected from document>",
        
        ...,

    ],

    "score": <number between -1 and 1>

}

\vspace*{\baselineskip}
\#\#\#Your Task\#\#\#

Selected sentences and score in dictionary form:

\end{tcolorbox}
\caption{}
\label{fig:ann_prompt_e}
\end{figure*}

\begin{figure*}
\begin{tcolorbox}[colback={gray!20},title={\textbf{Task prompt for pointwise evidence reranking: \stob and \btol }},colbacktitle=white,coltitle=black]
The following is a claim:
<claim>

A relevant passage provides supporting or refuting evidence for the claim.
\end{tcolorbox}
\caption{This task prompt takes the place of the query in the prompt used by the original Rank1 implementation.}
\label{fig:prompt-rank1-a}
\end{figure*}

\begin{figure*}
\begin{tcolorbox}[colback={gray!20},title={\textbf{Task prompt for pointwise evidence reranking: \stoe }},colbacktitle=white,coltitle=black]
I am writing an encyclopedia article about the following person: <entity>. A relevant passage contains noteworthy biographical facts about this person. For example, a passage containing facts about this person's early life, education, career, or death is relevant.
\end{tcolorbox}
\caption{Like the task prompt in \autoref{fig:prompt-rank1-a}, this task prompt takes the place of the query in the prompt used by the original Rank1 implementation.}
\label{fig:prompt-rank1-b}
\end{figure*}

\begin{figure*}
\begin{tcolorbox}[colback={gray!20},title={\textbf{Task prompt for listwise evidence reranking: \stob and \btol }},colbacktitle=white,coltitle=black]
The query given below is a claim. A relevant passage provides supporting or refuting evidence for the claim.
\end{tcolorbox}
\caption{This task prompt is prepended to the user message in the reranker's original implementation, which passes prompts to the model using the conversational format. The original system message, if it exists, is retained.}
\label{fig:prompt-listwise-a}
\end{figure*}

\begin{figure*}
\begin{tcolorbox}[colback={gray!20},title={\textbf{Task prompt for listwise evidence reranking: \stoe }},colbacktitle=white,coltitle=black]
I am writing an encyclopedia article about the following person given in the query below. A relevant passage contains noteworthy biographical facts about this person. For example, a passage containing facts about this person's early life, education, career, or death is relevant.
\end{tcolorbox}
\caption{Like the task prompt in \autoref{fig:prompt-listwise-a}, this task prompt is prepended to the user message in the reranker's original implementation, which passes prompts to the model using the conversational format. The original system message, if it exists, is retained.}
\label{fig:prompt-listwise-b}
\end{figure*}